\documentclass[hidelinks]{article}
\usepackage{natbib}
\usepackage{amssymb}
\usepackage[final]{nips_2017}
\usepackage[hidelinks]{hyperref}

\pdfoutput=1

\title{Multiset Canonical Correlation Analysis simply explained}

\author{
Lucas C. Parra\\
Department of Biomedical Engineering\\
City College of New York \\
\texttt{parra@ccny.cuny.edu}\\
} 

\begin{document}

\maketitle

\begin{abstract}%
There are a multitude of methods to perform multi-set canonical correlation analysis (MCCA), including some that require iterative solutions. The methods differ on the criterion they optimize and the constraints placed on the solutions.  This note focuses perhaps on the simplest version, which can be solved in a single step as the eigenvectors of matrix ${\bf D}^{-1} {\bf R}$. Here ${\bf R}$ is the covariance matrix of the concatenated data, and ${\bf D}$ is its block-diagonal. This note shows that this solution maximizes inter-set correlation (ISC) without further constraints. It also relates the solution to a two step procedure, which first whitens each dataset using PCA, and then performs an additional PCA on the concatenated and whitened data. Both these solutions are known, although a clear derivation and simple implementation are hard to find. This short note aims to remedy this.    
\end{abstract}

\section*{Background}

Canonical correlation analysis (CCA) was first introduced by \cite{hotelling1936relations} with the goal of identifying projections of two datasets that are maximally correlated. Hotteling's  approach was to maximize Pearson's correlation coefficient and to require that multiple projections be uncorrelated from one another. This formulation results in a closed-form solution based on the eigenvectors of a matrix suitably combining covariances of each data set and the cross-correlations between the two data sets. Hotteling's formulation is enduring because of it clarity and simplicity. 

Starting with Horst in 1961, a number of generalizations of this basic idea have been proposed for the case that there are more than two data sets \citep{horst1961generalized,horst1961relations,meredith1964rotation,carroll1968generalization}. These approaches were summarized for the first time in \citep{kettenring1971canonical}, which has become the standard reference for multi-set CCA (MCCA). Perhaps because Kettering offered 5 different optimization criteria for MCCA, the story did not end there. Since then, a number of additional methods have seen proposed by combining various optimality criteria with various constraints on the solutions \citep[no less than 20 versions are summarized by][Chapter 10]{asendorf2015informative}. There are additional, related methods and generalizations that were developed under different names, such as Generalized Factor Analysis \citep{kamronn2015multiview}, Generalized Procrustes Analysis \citep{gower1975generalized,gardner2006synthesis} or Hyperalignment \citep{haxby2011common}, which reinvents this concept decades later\citep{xu2012regularized}. 
The prolific multiplicity of approaches may be the result of the somewhat complex description of the methods in the early work.

\section*{Problem statement}

The purpose of this note is to give a short derivation for one of the simplest methods for MCCA. As with all MCCA methods, the goal here will be to find dimensions in multivariate data that maximize correlation between multiple data sets. Denote the data in the $l$th set with ${\bf x}_i^l \in \mathbb{R}^{d_l}$, where $i=1, \dots ,T$ enumerates exemplars, $l=1, \dots, N$ enumerates the data sets, and $d_l$ are the dimensions of each data set. In total the $N$ data sets have dimensions $D=\sum_{i=1}^N d_l$. The goal is to identify in each data set a linear combination of the $d_l$ measures, defined by a projection vectors ${\bf v}^l \in \mathbb{R}^{d_l}$:
\begin{equation}
y^l_i = {\bf v}^{l\top} {\bf x}^l_i , 
\label{projection}
\end{equation} 
such that the correlation of $y^l_i$ across exemplars $i$ between the $N$ data sets is maximized.  The present formulation defines the inter-set correlation (ISC) as the ratio of the between-set covariance, $r_B$, over the within-set covariance, $r_W$:
\begin{equation}
\rho = \frac{1}{N-1}\frac{r_B}{r_W} .
\label{isc}
\end{equation}
The between-set covariances are averaged over all pairs of sets, and within-set variances are averaged over all sets: 
\begin{eqnarray}
r_B &=& \sum_{l=1}^N \sum_{k=1, k \neq l}^N \sum_{i=1}^T   (y_i^l - {\bar y}_*^l)(y_i^k - {\bar y}_*^k) 
\label{between-set-covariance} , \\
r_W &=& \sum_{l=1}^N \sum_{i=1}^T (y_i^l - {\bar y}_*^l) (y_i^l - {\bar y}_*^l)
\label{within-set-covariance} ,
\end{eqnarray}
where, $\bar{y}^l_* =  \frac{1}{T} \sum_{i=1}^T y_i^l$, is the sample-mean for set $l$. The factor $(N-1)^{-1}$ in definition (\ref{isc}) is required so that correlation is normalized to $\rho \le 1$ \citep{parra2018correlated}. To simplify the notation, the definitions of ``variance'' and ``covariance'' omit the common scaling factor $(T-1)^{-1}N^{-1}$.

We have used this definition of correlation in the context of identifying maximal inter-subject correlation or inter-rater correlation using a method we called Correlated Component Analysis \citep[CorrCA,][]{parra2018correlated}. Here we deviate from this earlier work by allowing each data set to have its own projection vector ${\bf v}^l$.
By inserting Eq. (\ref{projection}) into definition (\ref{isc}) it follows readily that
\begin{equation}
\rho =  
\frac{
\sum_{l=1}^N \sum_{k=1, k \neq l}^N {\bf v}^{l\top} {\bf R}^{lk} {\bf v}^k
}
{
(N-1) \sum_{l=1}^N {\bf v}^{l\top} {\bf R}^{ll} {\bf v}^l
} ,
\label{isc-x}
\end{equation}
where ${\bf R}^{lk}$ are the cross-covariance matrices between ${\bf x}_i^l$ and ${\bf x}_i^k$:
\begin{equation}
{\bf R}^{lk} 
=
\sum_{i=1}^T 
({\bf x}_i^l - {\bar {\bf x}}_*^l)({\bf x}_i^k - {\bar {\bf x}}_*^k) ^\top ,
\end{equation}
Here, ${\bar {\bf x}}_*^l = \frac{1}{T} \sum_{i=1}^T {\bf x}_i^l$, is the sample-mean for data set $l$.

\section*{One-step solution}

The projection vectors ${\bf v}^l$ that maximize $\rho$ can be found as the solution of  $\partial \rho / \partial {\bf v}^{l\top} = 0$, which yields for each $l$ the following equation: 
\begin{equation}
\frac{1}{N-1} \sum_{k=1, k \neq l}^N {\bf R}^{lk} {\bf v}^k
=
{\bf R}^{ll} {\bf v}^l \rho  .
\end{equation}
This set of equations for all ${\bf v}^l$ can be written as a single equation if we concatenate them to a single ${\bf v} = \left[ {\bf v}^{1\top} \ldots {\bf v}^{1\top} \right]^\top$:
\begin{equation}
{\bf R} {\bf v}
=
{\bf D} {\bf v} \lambda,
\label{mcca-solution}
\end{equation}
where, $\lambda=(N-1)\rho+1$, ${\bf R}$ is a matrix combining all ${\bf R}^{lk}$, and the diagonal blocks ${\bf R}^{ll}$ are arranged in a block-diagonal matrix ${\bf D}$:

\begin{equation}
{\bf R} = 
\left[
\begin{array}{cccc}
{\bf R}^{11} & {\bf R}^{12} & \ldots & {\bf R}^{1N} \\
{\bf R}^{21} & {\bf R}^{22} & \ldots & {\bf R}^{2N} \\
\vdots & \vdots & \ddots & \vdots\\
{\bf R}^{N1} & {\bf R}^{N2} &\cdots & {\bf R}^{NN} \\
\end{array}
\right], 
{\bf D} = 
\left[
\begin{array}{cccc}
{\bf R}^{11} & 0 & \ldots & 0 \\
0 & {\bf R}^{22} & \ldots & 0 \\
\vdots & \vdots & \ddots & \vdots\\
0 & 0 &\cdots & {\bf R}^{NN} \\
\end{array}
\right].
\end{equation}
If ${\bf D}$ is invertible, the solution to the MCCA problem are simply the eigenvectors of ${\bf D}^{-1}{\bf R}$, or more generally, the eigenvectors of Eq. (\ref{mcca-solution}). We can arrange all such eigenvectors as columns to a single matrix ${\bf V} = [{\bf v}_1 \ldots {\bf v}_n \ldots  {\bf v}_D]$. The eigenvector with the largest eigenvalue $\lambda$ maximize the ISC because $\rho=\lambda-1$. The conventional constraint of the subsequent solutions is that they are uncorrelated from the first. This is satisfied by default as the solutions ${\bf V}$ to any eigenvalue equation of the form (\ref{mcca-solution}) diagonalize ${\bf D}$:
\begin{equation}
{\bf V}^T {\bf D} {\bf V} = {\bf \Lambda}.
\label{constraint}
\end{equation}  
And therefore each ${\bf v}^l_n$ generates a projection $y^l_n$ that is uncorrelated from all other $y^l_m, n \neq m$.

The solution (\ref	{mcca-solution}) to the MCCA problem is generally referred to as SUMCORR because it can be derived from maximizing the summed correlation subject to constraint (\ref{constraint}) with $ {\bf \Lambda} =  {\bf I}$ \citep[see, for instance,][]{nielsen2002multiset}. This MCCA solutions has also been the basis for various regularization methods, aiming at stabilizing the inverse ${\bf D}^{-1}$, many of them based on singular value decompositions of the data \citep{xu2012regularized,asendorf2015informative}. This is in particular important when ${\bf D}$ is rank deficient (many dimensions and few exemplars) \citep{fu2017scalable}.

\section*{Two-step solution}

As with any eigenvalue problem of a matrix ${\bf D}^{-1}{\bf R}$, with symmetric matrices ${\bf D}$ and ${\bf R}$, one can find the solutions also by executing two eigenvalue problems in sequence as follows. First, decompose ${\bf D}$ as
\begin{equation}
{\bf D} = {\bf U } {\bf \Lambda} {\bf U} 
\end{equation}
with orthonormal ${\bf U}$ and diagonal ${\bf \Lambda}$ (different from the one in Eq. \ref{constraint}).  Because ${\bf D}$ is the block-diagonal matrix of the covariances in each data set, this decomposition implies doing PCA on each data set separately, i.e whitening each data set. Then rewrite Eq. (\ref{mcca-solution}) as

\begin{eqnarray}
{\bf R} {\bf v} &=& {\bf U} {\bf \Lambda} {\bf U}^\top {\bf v} \lambda
\nonumber
\\
{\bf \Lambda}^{-1/2} {\bf U}^\top {\bf R} {\bf v} &=&  {\bf \Lambda}^{1/2} {\bf U}^\top {\bf v} \lambda
\nonumber
\\ 
{\bf \Lambda}^{-1/2} {\bf U}^\top {\bf R} {\bf U} {\bf \Lambda}^{-1/2} {\bf \Lambda}^{1/2} {\bf U}^\top  {\bf v} &=&  {\bf \Lambda}^{1/2} {\bf U}^\top {\bf v} \lambda 
\nonumber
\\
\tilde{{\bf R}} \tilde{{\bf v}} &=&  \tilde{{\bf v}} \lambda
\label{second-pca}
\end{eqnarray}
where, $\tilde{{\bf R}}={\bf \Lambda}^{-1/2} {\bf U}^\top {\bf R} {\bf U} {\bf \Lambda}^{-1/2}$, is nothing but the covariance of the whitened and concatenated data. Thus, the last eigenvalue equation (\ref{second-pca}) corresponds to performing PCA on the individually whitened and then concatenated data sets. The overall transformation on the data is then, ${\bf V} =  {\bf U} {\bf \Lambda}^{-1/2}\tilde{\bf V}$, with ${\bf U}$ and ${\bf \Lambda}$ being the eigenvectors and eigenvalues of the first PCA, and $\tilde{\bf V}$ the eigenvectors of the second PCA. 

\section*{Implementation}

Here is a short implementation of the one-step solution in matlab code. Note that the \verb|eig()| function typically sorts eigenvalues in ascending order. Therefore the last eigenvectors will give the component with the largest inter-set correlation $\rho$. In the case of two data sets, this code gives identical solutions to conventional CCA and $\rho$ is equal to Pearson's correlation coefficient.  

{
\begin{verbatim}
function [V,rho]=mcca(X,d)
% [V,rho]=mcca(X,d) Multiset Canonical Correlation Analysis. X is the data
% arranged as samples by dimension, whereby all sets are concatenated along
% the dimensions. d is a vector with the dimensions of each set. V are the
% component vectors and rho the resulting inter-set correlations.

N=length(d);
R=cov(X);
for i=N:-1:1, j=(1:d(i))+sum(d(1:i-1)); D(j,j)=R(j,j); end
[V,lambda]=eig(R,D);
rho = (diag(lambda)-1)/(N-1);
\end{verbatim}
}

\bibliography{mcca-references}

\begin{thebibliography}{}

\bibitem[Asendorf, 2015]{asendorf2015informative}
Asendorf, N.~A. (2015).
\newblock {\em Informative data fusion: Beyond canonical correlation analysis}.
\newblock PhD thesis, University of Michigan.

\bibitem[Carroll, 1968]{carroll1968generalization}
Carroll, J.~D. (1968).
\newblock Generalization of canonical correlation analysis to three or more
  sets of variables.
\newblock In {\em Proceedings of the 76th annual convention of the American
  Psychological Association}, volume~3, pages 227--228.

\bibitem[Fu et~al., 2017]{fu2017scalable}
Fu, X., Huang, K., Hong, M., Sidiropoulos, N.~D., and So, A. M.-C. (2017).
\newblock Scalable and flexible multiview max-var canonical correlation
  analysis.
\newblock {\em IEEE Transactions on Signal Processing}, 65(16):4150--4165.

\bibitem[Gardner et~al., 2006]{gardner2006synthesis}
Gardner, S., Gower, J.~C., and le~Roux, N.~J. (2006).
\newblock A synthesis of canonical variate analysis, generalised canonical
  correlation and procrustes analysis.
\newblock {\em Computational Statistics \& Data Analysis}, 50(1):107--134.

\bibitem[Gower, 1975]{gower1975generalized}
Gower, J.~C. (1975).
\newblock Generalized procrustes analysis.
\newblock {\em Psychometrika}, 40(1):33--51.

\bibitem[Haxby et~al., 2011]{haxby2011common}
Haxby, J.~V., Guntupalli, J.~S., Connolly, A.~C., Halchenko, Y.~O., Conroy,
  B.~R., Gobbini, M.~I., Hanke, M., and Ramadge, P.~J. (2011).
\newblock A common, high-dimensional model of the representational space in
  human ventral temporal cortex.
\newblock {\em Neuron}, 72(2):404--416.

\bibitem[Horst, 1961a]{horst1961generalized}
Horst, P. (1961a).
\newblock Generalized canonical correlations and their applications to
  experimental data.
\newblock {\em Journal of Clinical Psychology}, 17(4):331--347.

\bibitem[Horst, 1961b]{horst1961relations}
Horst, P. (1961b).
\newblock Relations amongm sets of measures.
\newblock {\em Psychometrika}, 26(2):129--149.

\bibitem[Hotelling, 1936]{hotelling1936relations}
Hotelling, H. (1936).
\newblock Relations between two sets of variates.
\newblock {\em Biometrika}, 28(3/4):321--377.

\bibitem[Kamronn et~al., 2015]{kamronn2015multiview}
Kamronn, S., Poulsen, A.~T., and Hansen, L.~K. (2015).
\newblock Multiview bayesian correlated component analysis.
\newblock {\em Neural Computation}, 27(10):2207--2230.

\bibitem[Kettenring, 1971]{kettenring1971canonical}
Kettenring, J.~R. (1971).
\newblock Canonical analysis of several sets of variables.
\newblock {\em Biometrika}, 58(3):433--451.

\bibitem[Meredith, 1964]{meredith1964rotation}
Meredith, W. (1964).
\newblock Rotation to achieve factorial invariance.
\newblock {\em Psychometrika}, 29(2):187--206.

\bibitem[Nielsen, 2002]{nielsen2002multiset}
Nielsen, A.~A. (2002).
\newblock Multiset canonical correlations analysis and multispectral, truly
  multitemporal remote sensing data.
\newblock {\em IEEE transactions on image processing}, 11(3):293--305.

\bibitem[Parra et~al., 2018]{parra2018correlated}
Parra, L.~C., Haufe, S., and Dmochowski, J.~P. (2018).
\newblock Correlated components analysis---extracting reliable dimensions in
  multivariate data.
\newblock {\em arXiv preprint arXiv:1801.08881}.

\bibitem[Xu et~al., 2012]{xu2012regularized}
Xu, H., Lorbert, A., Ramadge, P.~J., Guntupalli, J.~S., and Haxby, J.~V.
  (2012).
\newblock Regularized hyperalignment of multi-set fmri data.
\newblock In {\em Statistical Signal Processing Workshop (SSP), 2012 IEEE},
  pages 229--232. IEEE.

\end{thebibliography}

\end{document}